\documentclass[twocolumn]{article}

\usepackage[a4paper, margin=2cm]{geometry}

\usepackage{cite}
\usepackage{amsmath,amssymb,amsfonts}
\usepackage{graphicx}

\usepackage{bm}

\usepackage{xcolor}  
\usepackage{soul}  
\usepackage{bbm} 
\usepackage{adjustbox} 
\usepackage{comment}  
\usepackage[hidelinks]{hyperref}

\newcommand{\etal}{et~al.}
\newcommand{\ie}{i.e., }
\newcommand{\eg}{e.g., }

\newcommand\figureref[1]{Figure~\ref{#1}}
\newcommand\figurerefnotext[1]{\ref{#1}}
\newcommand\tableref[1]{Table~\ref{#1}}
\newcommand\sectionref[1]{Section~\ref{#1}}

\newcommand{\reducedmask}{\textit{reduced}}
\newcommand{\wroadagentsmask}{\textit{wroadagents}}

\newcommand{\wroadagentsnodrivablemask}{\textit{wroadagents\_nodrivable}}

\definecolor{darkblue}{RGB}{8, 69, 191}
\definecolor{darkred}{RGB}{181, 16, 16}

\DeclareMathOperator*{\argmax}{argmax}

\newcommand\Figure[3][width=\linewidth]{\begin{figure}\includegraphics[#1]{#2}\caption{#3}\end{figure}}

\begin{document}

\title{Learning Ordinality in Semantic Segmentation}

\author{Ricardo P. M. Cruz$^{1,2}$* \and Rafael Cristino$^1$ \and Jaime S. Cardoso$^{1,2}$*\\$^1$ Faculty of Engineering, University of Porto\\$^2$ INESC TEC, Porto\\ * rpcruz@fe.up.pt, jaime.cardoso@fe.up.pt}
\date{\textbf{Paper published in IEEE Access:} \href{http://doi.org/10.1109/ACCESS.2025.3537601}{10.1109/ACCESS.2025.3537601}}

\maketitle

\begin{abstract}
Semantic segmentation consists of predicting a semantic label for each image pixel.
While existing deep learning approaches achieve high accuracy, they often overlook the ordinal relationships between classes, which can provide critical domain knowledge (e.g., the pupil lies within the iris, and lane markings are part of the road).
This paper introduces novel methods for \emph{spatial ordinal segmentation} that explicitly incorporate these inter-class dependencies. By treating each pixel as part of a structured image space rather than as an independent observation, we propose two regularization terms and a new metric to enforce ordinal consistency between neighboring pixels.
Two loss regularization terms and one metric are proposed for structural ordinal segmentation, which penalizes predictions of non-ordinal adjacent classes.
Five biomedical datasets and multiple configurations of autonomous driving datasets demonstrate the efficacy of the proposed methods. Our approach achieves improvements in ordinal metrics and enhances generalization, with up to a 15.7\% relative increase in the Dice coefficient. Importantly, these benefits come without additional inference time costs. This work highlights the significance of spatial ordinal relationships in semantic segmentation and provides a foundation for further exploration in structured image representations.
\end{abstract}

\section{Introduction}

Semantic segmentation, or scene parsing, is the task of attributing a semantic label to each of the pixels in an image, resulting in a segmentation map. One common problem with these deep learning segmentation models is the lack of generalization ability, which means that the network fails to make appropriate predictions when parsing a situation that did not occur in the training dataset~\cite{zakaria2023lane,yousri2021deep}. While convolutional neural networks are designed to first capture low-level features from nearby pixels and progressively learn more abstract, high-level features from distant pixels~\cite{zeiler2014visualizing}, high-level relations are not explicitly specified. A hypothesis is that the neural network model does not have the necessary intrinsic domain knowledge of the task -- it may fail to infer appropriate high-level relations from the data used to train it (e.g., the pupil being inside the iris in an eye). 

In many situations, there is an explicit ordering between the output classes, and by training the network with methods that uphold the ordinal constraints, the network may be able to learn better higher-level concepts, such as the ordinal relation between different objects (\eg the lane marks and are inside the lane, etc) and the relative placement of objects (\eg the sidewalk is to the side of the road, etc.).

Typically, ordinal problems have mostly been studied in the context of (image) classification~\cite{cruz2017ordinal,albuquerque_ordinal_2021,albuquerque2022quasi,bellmann2020ordinal}, where the task is to classify an observation (image) as one of $\mathcal C_1 \prec \mathcal C_2 \prec \ldots \prec \mathcal C_K$ ordered classes (for example, the severity of a disease), as opposed to nominal classes in the case of classic nominal classification. DORN~\cite{fu2018deep} has proposed using ordinal losses to promote order at the pixel level in-depth estimation to avoid inconsistencies where a given pixel is classified as having a high probability of being near and also a high probability of being far away but a low probability of being at a medium distance. But, to our knowledge, only one work~\cite{fernandes_ordinal_2018} attempted to introduce ordinal relations between parts of the segmentation, where $\mathcal C_1 \supset \mathcal C_2 \supset \ldots \supset \mathcal C_K$. However, the work does not explicitly perform spatial segmentation: it would be expected that an area labeled as \(\mathcal C_k\) would only have a direct boundary to the areas segmented as \(\mathcal C_{k-1}\) and \(\mathcal C_{k+1}\), as exemplified in \figureref{fig:ordseg_viz_example}.

\Figure{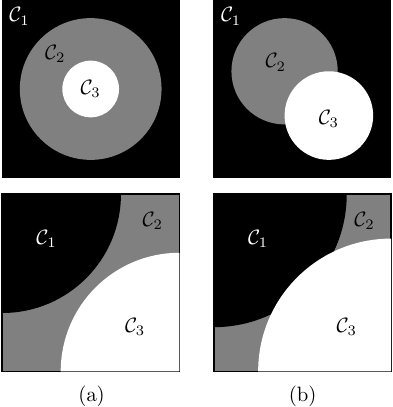}{\textbf{Example of (a)~segmentation masks and (b)~hypothetical non-constrained model outputs for an ordinal problem with three distinct classes, $\{\mathcal C_1,\mathcal C_2,\mathcal C_3\}$, where such an order is defined that \(\mathcal C_1 \supset \mathcal C_2 \supset \mathcal C_3\), therefore, an area segmented as \(\mathcal C_1\) can only possibly have a direct boundary with areas segmented as \(\mathcal C_2\), whereas \(\mathcal C_2\) can have boundaries both with \(\mathcal C_1\) and \(\mathcal C_3\).}\label{fig:ordseg_viz_example}}

The transition from image classification to segmentation inherently evolves the prediction task from classifying independent samples (each image) to classifying structured dependent samples (each pixel in an image). The former can only take action in the representation space of each pixel, while the latter can also consider the structured space and the relations between the samples. Further developing this idea, the structured image space can be generalized to a graph, where each pixel is a vertex and is connected to its adjacent pixels.

While ordinal consistency has been little explored, \cite{melo2023retinal} has proposed using different models for segmentation masks of classes contained inside each other. Other types of consistency from the literature include hierarchical segmentation~\cite{li2022deep}, where each pixel has a taxonomy (e.g., a pixel is segmented as a small/large vehicle and then further segmented as motorcycle/bicycle if small or car/bus/truck if large). Furthermore, spatial consistency has been explored in terms of part segmentation -- super-pixels have been used to impose constraints, such as the ``head'' appearing above ``upper body'' or ``hair'' being above ``head''~\cite{bo2011shape}, and similarly, the function of Restricted Boltzmann Machines has been used to favor certain classes based on their spatial locations~\cite{tsogkas2015deep}. Recurrent modules have also been used to iteratively predict segmentation parts by establishing a predefined order between the parts~\cite{zhao2019ordinal}. Neither of these relates to ordinal segmentation in that sub-segmentations are contained inside super-segmentations.

The present work extends the dissertation~\cite{cristino2023introducing} with the following contributions:
\begin{itemize}
\item The task of ordinal segmentation is formalized in two consistency tasks: (i)~representation consistency focuses on ordinal consistency within each individual pixel, and (ii)~structural consistency focuses on ordinal consistency between structures of the input. 
\item Focusing on the latter case of structural consistency, for which literature is lacking, two loss regularization terms are proposed, as well as a novel metric. 
\item A thorough evaluation is performed for five biomedical datasets and an autonomous driving dataset. Methods that improve metrics of ordinal consistency are shown to also improve the Dice coefficient (a common non-ordinal semantic segmentation metric) up to 15.7\% in relative terms.
\end{itemize}

The document is structured as follows: \sectionref{sec:sota} introduces the state of the art; \sectionref{sec:proposal} delineates the proposal; \sectionref{sec:experiments} describes the conducted experiments; \sectionref{sec:results} shows and analyzes the experimental results; and \sectionref{sec:conclusion} revisits the proposed methods and summarizes the key conclusions.


\section{State of the Art}
\label{sec:sota}

Semantic segmentation, the task of assigning pixel-level labels to an image, has seen significant advancements due to the rise of convolutional neural networks. Early approaches, such as the Fully Convolutional Network (FCN)~\cite{long2015fully}, replaced traditional fully connected layers with convolutional layers to enable dense pixel-wise predictions. This led to architectures like U-Net~\cite{ronneberger_u-net_2015} and SegNet~\cite{badrinarayanan2017segnet} that incorporated encoder-decoder structures to capture both high-level semantics and fine-grained details.
Further refinements include the use of dilated (or atrous) convolutions, as introduced in DeepLab~\cite{chen2018encoder,das2021estimation}, to capture multi-scale context without losing resolution. Another development is the incorporation of attention mechanisms, such as in PSPNet~\cite{zhao2017pyramid} and the more recent Transformer-based approaches, e.g. ones based on Swin~\cite{zongren2023densetrans}, which have improved performance by better capturing long-range dependencies.

\subsection{Loss Functions}

In these works, a common loss is cross entropy, albeit auxiliary losses may also be used, such as a differentiable version of the Dice coefficient. Defining cross entropy for a semantic segmentation problem,
\begin{equation}
\label{equation:ce}
\mathrm{CE}(\bm{y}_n, \hat{\bm{p}}_n) = - \frac{1}{HW} \sum_{i=1}^H \sum_{j=1}^W \sum_{k=1}^K \mathbbm{1}(y_{n,i,j}=k) \log(\hat{p}_{n,k,i,j}),
\end{equation}
where $\hat{\bm{p}}$ is the model output as probabilities, in shape $(N, K, H, W)$, where $N$ is the batch size, $K$ is the number of classes, and $(H, W)$ are, respectively, the height and width of each image; $\bm{y}$ is the ground truth segmentation map, in shape $(N, H, W)$, where each value $y_{n,i,j}$ corresponds to the ground truth class $k \in \{1,\cdots, K\}$ of the pixel at position $(i,j)$ of observation $n$; and $\mathbbm{1}(x)$ is the indicator function of $x$.

Cross-entropy maximizes the probability of the ground truth class for each pixel in the observation, ignoring the distribution of the predictions for the other classes. This is a potential area where new loss functions can improve by restricting the probabilities of the non-ground truth class according to the domain knowledge of the task.

\subsection{Ordinal Classification Methods} \label{sec:sota:ordinal_methods}

Various research works seek to imbue deep neural networks with ordinal domain knowledge in the ordinal classification domain.

\subsubsection{Unimodality} \label{sec:sota:ordinal_methods:co2}

The promotion of unimodality in the distribution of the model output probabilities has achieved good results in ordinal classification tasks~\cite{pinto_da_costa_unimodal_2008,beckham_unimodal_2017,albuquerque_ordinal_2021,cardoso_unimodal_2023}. This can be advantageous in ordinal problems because the model should be more uncertain between ordinally adjacent classes. For example, it would not make sense for a model to output a high probability for Low and High risk of disease but a small probability for Medium risk of disease. \figureref{fig:unimodality_pixel} shows the difference between multimodal and unimodal distributions.

\Figure{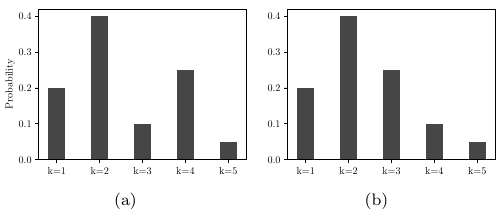}{\textbf{Example of possible (a)~multimodal and (b)~unimodal output probability distributions for a given pixel.}\label{fig:unimodality_pixel}}

To promote unimodal output probability distributions, some authors have imposed architectural restrictions, which restrict the network output to a Binomial or Poisson probability distributions~\cite{pinto_da_costa_unimodal_2008,beckham_unimodal_2017}. Other authors have promoted ordinality by penalizing the model when it outputs a non-unimodal distribution through augmented loss functions. One such case is the CO2 loss, which penalizes neighbor class probabilities if they do not follow unimodal consistency~\cite{albuquerque_ordinal_2021},
\begin{equation}
\label{equation:classification_co2}
L_\mathrm{CO2}(y_n, \hat{\bm{p}}_n) = L_\mathrm{CE}(y_n, \hat{\bm{p}}_n) 
+ \lambda  L_\mathrm{O2}(y_n, \hat{\bm{y}}_n),
\end{equation}
where $L_\text{O2}$ is the regularization term,
\begin{align}
\begin{split}
\label{equation:classification_co2_term}
L_\mathrm{O2}(y_n, \hat{\bm{p}}_n) &= \sum_{k=1}^{K-1} \mathbbm{1}(k \geq y_{n}) \,\mathrm{ReLU}(\delta + \hat{p}_{n,k+1} - \hat{p}_{n,k}) \\
&+ \sum_{k=1}^{K-1} \mathbbm{1}(k \leq y_{n}) \,\mathrm{ReLU}(\delta + \hat{p}_{n,k} - \hat{p}_{n,k+1}),
\end{split}
\end{align}
with $\delta$ being an imposed margin, assuring that the difference between consecutive probabilities is at least $\delta$, and ReLU is defined as $\mathrm{ReLU}(x) = \max(0, x)$.

\subsubsection{Ordinal Encoding}

An approach to introducing ordinality to neural networks for classification involves regularizing the input data by encoding the ordinal distribution in the ground truth labels~\cite{cheng_neural_2007}. Defining $k^\star$ as the ground truth class for a given sample, this input data encoding encodes each class as $\mathbbm{1}(k < k^\star)$, whereas generic one-hot encoding encodes each class as $\mathbbm{1}(k = k^\star)$.
This approach can be adapted for segmentation problems by similarly encoding the ground truth masks at a pixel level~\cite{fernandes_ordinal_2018}. 

Using ordinal encoding for segmentation does not guarantee that the output probabilities are monotonous, \ie the probability of ordinal class $k$, $P_k$, may be less than $P_{k+1}$. The consistency of the output class probabilities can be achieved by using,
\begin{equation}
\label{equation:pixelwise_consistency}
\text P(\mathcal C^+_{k+1}) = \text P(\mathcal C^+_{k+1} \mid \mathcal C_k^+) \text P(\mathcal C_k^+),
\end{equation}  
where $\text P(\mathcal C^+_{k+1} \mid \mathcal C_k^+)$ is the $(k+1)$-th output of the network and $\text P(\mathcal C_k^+)$ is the corrected probability of class $k$~\cite{fernandes_ordinal_2018}. 

\bigskip
The state-of-the-art ordinal segmentation approaches have treated pixels as independent observations and promoted ordinality in their representation. However, this may be insufficient when applied to the structured image space, where pixels are dependent observations.


\section{Proposal}
\label{sec:proposal}

Firstly, we present the formal foundation of our work (\sectionref{sec:foundation}), then introduce the proposed ordinal segmentation methods, which are categorized into representation consistency (\sectionref{sec:proposal:pixel_wise}), and structural consistency (\sectionref{sec:proposal:spatial}). Finally, the ordinal segmentation problem is adapted to domains with arbitrary hierarchies (\sectionref{sec:proposal:hierarchies}).

\subsection{Foundation}
\label{sec:foundation}

We start by recovering the definition of ordinal models, as introduced in \cite{cardoso2010classification,sousa2011ensemble}.
In a model consistent with the ordinal setting, a small change in the input data should not lead to a ``big jump'' in the output decision. Assuming $f(\mathbf{x})$ as a decision rule that assigns each input value $\mathbf{x}\in \mathbb R^d$ to the index $\in\{1,2,\dots,K\}$ of the predicted class, the decision rule is said to be consistent with an ordinal data classification setting in a point $\mathbf x_0$ only if
$$\exists~\varepsilon>0, \ \max_{\mathbf x\in\mathcal B_\varepsilon(\mathbf x_0)} f(\mathbf x)-\min_{\mathbf x\in\mathcal B_\varepsilon(\mathbf x_0)} f(\mathbf x)\leq1,$$
with $\mathcal B_\varepsilon$ representing the individual feature-space neighborhood centered in $\mathbf{x}_0$ with radius $\varepsilon$. Equivalently, the decision boundaries in the input feature-space $\mathbf x$ should be only between regions of consecutive classes. Note that the concept of consistency with the ordinal setting is independent of the model type (probabilistic or not) and relies only on the decision region produced by the model. The state-of-the-art methods discussed in the previous Section focus only on this consistency (albeit often indirectly, by working on a related property, like the unimodality in the output probability space).

\Figure{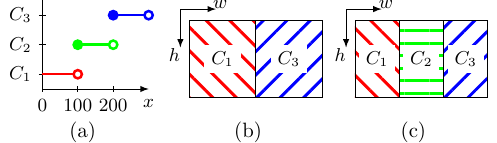}{\textbf{Illustration of possible ordinal (in)consistencies. (a)~Ordinal representation consistency. (b)~Ordinal structure inconsistency. (c)~Ordinal structure consistency.}\label{fig:ordinal-consistency}}

However, this consistency is only part of our knowledge in the ordinal segmentation setting.
In \figureref{fig:ordinal-consistency}, the feature description ${\mathbf x} \in {\mathbb R}$ at every pixel is mapped to the class according to the representation function
\begin{align}
f(x)=
\begin{cases}
\mathcal C_1, \text{ if } \quad x<100\\
\mathcal C_2, \text{ if } \quad 100\leq x<200\\
\mathcal C_3, \text{ if } \quad x\geq 200,
\end{cases}
\end{align}
which is also depicted in \figureref{fig:ordinal-consistency}(a). Note that the model is ordinal-consistent in the pixel description space.
At most, a small change in $x$ will change the decision to an adjacent class.

Consider now that the representation learned at each pixel $v$ is given as 
\begin{align}
\mathbf{x}(v)=
\begin{cases}
\mathbf{x}(v_0),  & \text{ if } \quad {v} \in \text{ left half of the image }\\
4\mathbf{x}(v_0), & \text{ if otherwise,}
\end{cases}
\end{align}
where $\mathbf{x}(v_0)$ is arbitrarily defined.
The model always sets the right half of the image to four times the values in the left half. Therefore, as illustrated in \figureref{fig:ordinal-consistency}(b), even being consistent at the pixel representation level, for some images (e.g., when the description of the pixels on the left takes the value 75), the decision will be $\mathcal C_1$ for the left half and $\mathcal C_3$ for the right half, which is to be avoided in an ordinal segmentation scenario. Therefore, we generalize the definition of ordinal-consistent models to models acting on structured observations, such as images or, more generally, graphs.

Consider each image as a graph $ \mathcal{G}=(\mathcal{V}, \mathcal{E}) $, where $\mathcal{V}$ denotes the set of graph vertices and $\mathcal{E}\subset \mathcal{V}\times \mathcal{V}$ denotes the set of graph edges. 
Vertices in the image graph represent pixels, and because the values of a pixel are usually highly related to the values of its neighbors, there are undirected edges from a pixel to its neighboring pixels (often 4 or 8). The corresponding graph is then a 2D lattice.
Consider the goal of learning a function of signals/features which takes as input:
\begin{itemize}
\item A feature description $\mathbf{x}_i = g(v_i)$ for every node $v_i$
\item A representative description of the graph structure $\mathcal{G}=(\mathcal{V}, \mathcal{E})$
\end{itemize}
and produces a node-level output $f(\mathbf{x}_i)$.
Also, remember that the closed neighborhood of a vertex $v$ in a graph $ \mathcal{G}$ is the subgraph of $\mathcal{G}$, $\mathcal{N}[v]$, induced by all vertices adjacent to $v$ and $v$ itself, i.e., the graph composed of the vertices adjacent to $v$ and $v$, and all edges connecting vertices adjacent to $v$.

We now define a model as ordinal consistent if the following two conditions are simultaneously met:
\begin{itemize}
\item {\bf representation consistency}: as before, if, for every point $\mathbf{x}_0 \in \mathbb{R}^d$, $\exists~\varepsilon>0$, $\max_{\mathbf x\in\mathcal B_\varepsilon(\mathbf x_0)} f(\mathbf x)-\min_{\mathbf x\in\mathcal B_\varepsilon(\mathbf x_0)} f(\mathbf x)\leq1$, with $\mathcal B_\varepsilon$ representing the feature-space neighborhood centered in $\mathbf{x}_0$ with radius $\varepsilon$.
\item {\bf structural consistency}: if, for every node $v_i$,\\ $\max_{v\in {\cal N}[v_i]} f(g(v))-\min_{v\in {\cal N}[v_i]} f(g(v)) \leq 1$
\end{itemize}
Figure~\ref{fig:ordinal-consistency}(b) illustrates structural inconsistency and consistency -- the representations of the pixels change smoothly over the image as given by
\begin{align}
\mathbf{x}(v)=
\begin{cases}
\mathbf{x}(v_0),     & \text{ if } v \in \text{left third of the image }\\
\mathbf{x}(v_0) + 1, & \text{ if } v \in \text{middle third of the image }\\
\mathbf{x}(v_0) + 2, & \text{ if } v \in \text{right third of the image.}
\end{cases}
\end{align}

\subsection{Representation Consistency for Ordinal Segmentation} \label{sec:proposal:pixel_wise}

In ordinal segmentation, ordinal representation consistency methods encompass those methods that act on the individual pixel representation $\mathbf{x}(v)$ and decision, \ie they impose restrictions on the pixel, taking into account its own characteristics and disregarding the context of the neighboring pixels. Such methods include the ordinal pixel encoding and consistency methods discussed in the state-of-the-art analysis~\cite{fernandes_ordinal_2018}. The previously introduced O2 term \eqref{equation:pixelwise_consistency} can be trivially adapted to segmentation by performing the regularization term for each pixel,
\begin{align}
\begin{split}
\label{equation:segco2_term}
L_\text{O2}(\bm{y}_n, \hat{\bm{p}}_n) = &\frac{1}{HW}\sum_{i=1}^W\sum_{j=1}^H
\Bigg[ \\
&\sum_{k,k'\in\mathcal S^A(y_{n,i,j})} \mathrm{ReLU}(\delta + \hat{p}_{n,k,i,j} - \hat{p}_{n,k',i,j}) + \\
& \sum_{k,k'\in\mathcal S^D(y_{n,i,j})} \mathrm{ReLU}(\delta + \hat{p}_{n,k,i,j} - \hat{p}_{n,k',i,j}) \Bigg].
\end{split}
\end{align}
where $k,k'\in\mathcal S^A(y_{n,i,j})$ and $k,k'\in\mathcal S^D(y_{n,i,j})$ are pairs of classes that have an ascending and descending order, respectively, which typically would be $S^A(y_{n,i,j})=\{(k,k+1)\mid k<y_{n,i,j}\}$ and $S^D(y_{n,i,j})=\{(k+1,k)\mid k>y_{n,i,j}\}$.

\subsection{Structural Consistency for Ordinal Segmentation} \label{sec:proposal:spatial}

The structural neighborhood concerns the neighborhood that relates one element with the others in the graph structure. In the case of semantic segmentation, the structural neighborhood is the set that contains all pixels connected to a given pixel. 

\subsubsection{Contact Surface Loss Using Neighbor Pixels} \label{chap:main:csnp}

\newcommand{\csnploss}{\mathrm{L_{CSNP}}}
\newcommand{\csnpterm}{\mathrm{CSNP}}
\newcommand{\ijsum}{\sum_{i=1}^H \sum_{j=1}^W}

The contact surface loss is defined by penalizing the prediction of two neighboring pixels of non-ordinally adjacent classes. This penalization term is performed by multiplying the output probabilities for neighboring pixels,
\begin{align}
\begin{split}
\label{equation:cs_tv}
L_\text{CSNP}(\hat{\bm{p}}_n,\hat{\bm{p}}_m) = \sum_{\substack{n,m : \\v_n\in {\cal N} [v_m]}}   \hat{\bm{p}}_n^\intercal C \hat{\bm{p}}_m
\end{split}
\end{align}
where $\hat{\bm{p}}_n^\intercal C \hat{\bm{p}}_n$ is a symmetric bilinear form, $C$ is a $K\times K$ cost matrix, $\hat{\bm{p}}_n\in\mathbb R^K$ is the vector of probabilities predicted at node $v_n$ (similarly for $\mathbf{p}_{m}$). $C$ is defined as
\begin{equation}
C_{i,j} = ReLU (|i-j| - 1) = 
\begin{cases}
0, & \text{if } |i-j|\leq 1 \\
C_{i,j} = |i-j|-1,  & \text{otherwise,}
\end{cases}
\end{equation}
to penalize ordinal-inconsistent neighboring probabilities,
where $\text{ReLU}(\cdot)$ is the Rectified Linear function, $\text{ReLU}(x) = \max(0,x)$. As an example, for $K=4$ classes, we obtain 
$$
\setlength\arraycolsep{3pt}
C= \begin{bmatrix}
0 & 0 & 1 & 2\\
0 & 0 & 0 & 1\\
1 & 0 & 0 & 0\\
2 & 1 & 0 & 0 
\end{bmatrix}.
$$
The intuition is to penalize high probabilities in spatially-close classes that are not adjacent.
%

\subsubsection{Contact Surface Loss Using the Distance Transform} \label{chap:main:csdt}

\newcommand{\DThead}{
    \mathrm{DT}(\hat{\bm{p}}_{n,k})_{p_1}
}

Another approach is leveraging the distance transform, which is an image map where each value represents the distance from each pixel in the target image to its closest pixel with value 1, calculated with the customizable distance function $d$~\cite{strutz2021distance}. Defining the distance transform (DT) of the output probability map of class $k$, 
\begin{equation}
\label{equation:dt_k}
\DThead = \min_{p_2 : \hat{p}_{n,k,p_2} \geq \delta} d(p_1, p_2),
\end{equation}
where $p = (i,j)$ and $\delta$ is the threshold parameter that selects the high-confidence pixels (typically 0.5), allowing the distance transform to be calculated for a high-certainty version of the output segmentation mask.

\newcommand{\csdtloss}{\mathrm{L_{CSDT}}}
\newcommand{\csdtterm}{\mathrm{CSDT}}

This provides a distance between pixels that have a high probability for a certain class and the closest pixel with a low probability of that class. That is, it provides an approximate distance between pixels of different classes. The model is trained to maximize this distance by multiplying the class probabilities map with the opposing class's distance transform,
\begin{equation}
\label{equation:cs_dt}
\begin{split}
L_\text{CSDT}(\hat{\bm{p}}_n) = -\frac{1}{|\mathcal S|} \sum_{k_1, k_2 \in \mathcal S} C_{k_1, k_2}
&\big(\hat{\bm{p}}_{n,k_1} \mathrm{DT}(\hat{\bm{p}}_{n,k_2}) \\ & + \hat{\bm{p}}_{n,k_2} \mathrm{DT}(\hat{\bm{p}}_{n,k_1})\big),
\end{split}
\end{equation}
where $\mathcal S$ is the set of pairs $(k_1, k_2) \in \{1,\cdots, K\}$ such that $k_2-k_1 > 1$.

At this stage, the $\csdtterm$ term maximizes the distance indefinitely. This is problematic because this may cause exploding distances, drawing the masks away from each other and possibly completely deviating from the ground truth. This can be solved by limiting the distance transform to a maximum distance, $\gamma$. This way, the loss only penalizes masks closer to each other than the $\gamma$ value. For that reason, an updated distance transform was used, which saturates,
\begin{equation}
\label{equation:dt_k:limited}
\DThead^{\gamma} = \min (\DThead, \gamma).
\end{equation}

\subsection{Partially Ordered Domains} \label{sec:proposal:hierarchies}

The previous ordinal segmentation methods were proposed for ordinal domains with a (linear) total order in the set of classes, \ie domains where $\mathcal C_1 \supset \mathcal C_2 \supset \ldots \supset \mathcal C_K$. This section proposes extending the previous work on ordinal losses to partially ordered output sets (\eg the road may contain both lane marks and vehicles, but there is no relation between lane marks and vehicles).

\begin{figure}
\centering
\includegraphics{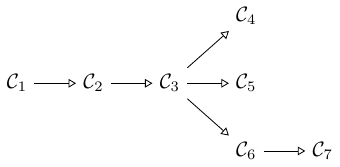}
\caption{A Hasse diagram exemplifying a domain with a partial order in its set of classes.}
\label{fig:ordseg:reduced1:hassediagram}
\end{figure}

Domains with a partial order in their set of classes must provide the segmentation methods with this partial order, \ie the ordinal relations that motivate the ordinal constraints, as exemplified by Figure~\ref{fig:ordseg:reduced1:hassediagram}.

The ordinal segmentation metrics and losses proposed to total order sets can be extended to partially ordered sets.
Define first ${\ell}_{m,n}$ as the length of the shortest path from class $m$ to class $n$ in the Hasse diagram, or $0$ if no such path exists. Note that ${\ell}_{m,n} \not = {\ell}_{n,m}$ and ${\ell}_{m,n} \vee {\ell}_{n,m}  = 0$.

The previously introduced loss terms ($L_\text{O2}$, $L_\csnpterm$ and $L_\csdtterm$) are extended to arbitrary hierarchies as $L_\text{O2}'$, $L_\csnpterm'$ and $L_\csdtterm'$.

The $L_\mathrm{O2}'$ term can be redefined from \eqref{equation:segco2_term} by applying set $\mathcal S$ between two neighboring classes whenever ${\ell}_{m,n} = 1$, and if the edge $(m,n)$ is part of a Hasse path that includes the ground-truth label $k^\star$.

The $C$ cost matrix in the $L_\csnpterm'$ term can be defined as 
$$C_{m,n} = \text{ReLU}(\max(\ell_{m,n},\ell_{n,m})-1).$$ 

The adaptation $L_\csdtterm'$ is similar to the adopted in $L_\mathrm{O2}'$. 

Note that all these extensions revert to the base versions when applied to a totally ordered set.

\subsection{Evaluation Metrics}
\label{sec:proposal:metrics}
\subsubsection{Unimodal Pixels}
We propose measuring the representation consistency using as a metric the percentage of Unimodal Pixels (UP), originally proposed by~\cite{cardoso_unimodal_2023}, which consists of the fraction of times that the probability distribution produced by the model is unimodal.
\subsubsection{Contact Surface Metric}
To evaluate the structural consistency, we propose to measure the percentage of ordinally invalid inter-class jumps between adjacent pixels, a metric for the contact surface between the masks of non-ordinally adjacent classes. Ordinally valid jumps are considered to be jumps between classes whose ordinal distance equals 1. If the ordinal distance between the classes of adjacent pixels exceeds 1, then that is an ordinally invalid jump. This requires that each pixel and its immediate neighborhood be examined during calculation. Defining the Contact Surface (CS) metric,
\begin{equation}
\label{equation:csmetric}
\begin{split}
\mathrm{CS}(\hat{\bm{y}}_n) &= \frac{1}{2}\frac{\ijsum \mathbbm{1} (\mathrm{CS_{dx}}(\hat{\bm{y}}_n)_{i,j} \geq 2)}{\ijsum \mathbbm{1} (\mathrm{CS_{dx}}(\hat{\bm{y}}_n)_{i,j} \geq 1)} \\&+ \frac{1}{2}\frac{\ijsum \mathbbm{1} (\mathrm{CS_{dy}}(\hat{\bm{y}}_n)_{i,j} \geq 2)}{\ijsum \mathbbm{1} (\mathrm{CS_{dy}}(\hat{\bm{y}}_n)_{i,j} \geq 1)},
\end{split}
\end{equation}
where $\hat{\bm{y}} = \argmax_{k=1}^K(\hat{\bm{p}})$, and $\mathrm{CS_{dx}}$ and $\mathrm{CS_{dy}}$ are the ordinal index variation, \ie ordinal distance, from the current pixel $(i, j)$ to the neighborhood, respectively, through the $x$ and $y$ axis,
\begin{equation}
\begin{split}
\mathrm{CS_{dx}}(\hat{\bm{y}}_n)_{i,j} = |\hat{y}_{n,i, j} - \hat{y}_{n,i, j+1}|\\
\mathrm{CS_{dy}}(\hat{\bm{y}}_n)_{i,j} = |\hat{y}_{n,i, j}  - \hat{y}_{n,i+1, j}|
\end{split}
\end{equation}


\section{Experiments} \label{sec:experiments}

\subsection{Datasets}
\label{sec:experiments:datasets}

Various real-life biomedical datasets with ordinal segmentation tasks, \ie where there is a clear ordering between classes, were identified from the literature~\cite{fernandes_ordinal_2018}. \tableref{table:sota:ordinal_datasets} introduces the five biomedical datasets used to validate the proposed methods, along with a sample image and its corresponding segmentation mask.

\begin{table}
\setlength{\tabcolsep}{0pt}  
\centering
\caption[A selection of appropriate biomedical datasets for ordinal segmentation]{A selection of appropriate biomedical datasets for ordinal segmentation.}
\label{table:sota:ordinal_datasets}
\begin{tabular}{|l|c|c|c|}
\hline
Dataset & \# Images & \# Classes & Sample \\
\hline
Breast Aesthetics~\cite{cardoso_towards_2007} & 120 & 4 & \raisebox{-0.4\totalheight}{\includegraphics[height=6ex]{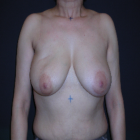}}\hspace{1mm}\raisebox{-0.4\totalheight}{\includegraphics[height=6ex]{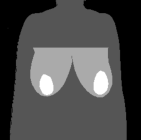}} \\
Cervix-MobileODT~\cite{noauthor_intel_nodate} & 1480 & 5 & \raisebox{-0.4\totalheight}{\includegraphics[height=6ex]{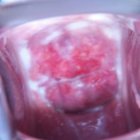}}\hspace{1mm}\raisebox{-0.4\totalheight}{\includegraphics[height=6ex]{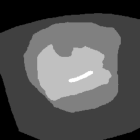}} \\
Mobbio~\cite{sequeira_mobbio_2014} & 1817 & 4 & \raisebox{-0.4\totalheight}{\includegraphics[height=6ex]{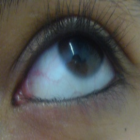}}\hspace{1mm}\raisebox{-0.4\totalheight}{\includegraphics[height=6ex]{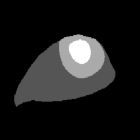}} \\
Teeth-ISBI~\cite{wang_benchmark_2016} & 40 & 5 & \raisebox{-0.4\totalheight}{\includegraphics[height=6ex]{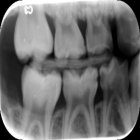}}\hspace{1mm}\raisebox{-0.4\totalheight}{\includegraphics[height=6ex]{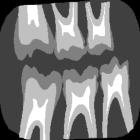}} \\
Teeth-UCV~\cite{fernandez_teethpalate_2012} & 100 & 4 & \raisebox{-0.4\totalheight}{\includegraphics[height=6ex]{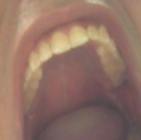}}\hspace{1mm}\raisebox{-0.4\totalheight}{\includegraphics[height=6ex]{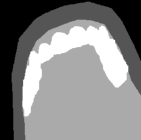}}\\
\hline
\end{tabular}
\end{table}

To evaluate the ordinal methods on autonomous driving domains, the BDD100K~\cite{bdd100k} and Cityscapes~\cite{cordts_cityscapes_2016} datasets were used:
\begin{itemize}
\item BDD100K -- is a multi-task, large-scale, and diverse dataset, obtained in a crowd-sourcing manner. Its images are split into two sets, each supporting a different subset of tasks: (1)~100K images -- 100,000 images with labels for the object detection, drivable area, and lane marking tasks, and (2)~10K images -- 10,000 images with labels for the semantic segmentation, instance segmentation, and panoptic segmentation tasks. The 10K dataset is not a subset of the 100K, but considerable overlap exists.
\item Cityscapes -- is a large-scale and diverse dataset, with scenes obtained from 50 different cities. It provides 5,000 finely annotated images for semantic segmentation.
\end{itemize}

Two variants of the BDD100K dataset were considered: (1)~BDD10K for the ordinary semantic segmentation task (10,000 images); and (2)~BDDIntersected, which is the intersection of the 100K and 10K subsets and supports both the semantic segmentation and drivable area tasks (2,976 images). The models trained with BDD10K were subsequently tested with Cityscapes to validate the methods' generalization ability with out-of-distribution (OOD) testing. Furthermore, to evaluate how the proposed methods influence learning with scarce data, a dataset scale variation experiment was conducted with the BDD10K dataset.

To transpose semantic segmentation in autonomous driving to an ordinal segmentation problem, ordinal relations must be derived from the classes in the dataset. When analyzing an autonomous driving scene, \eg \figureref{fig:driving_scene1_bdd100k}, we can, a priori, derive that, usually:

\begin{itemize}
\item The vehicles will be on the road or in parking spaces;
\item The drivable area will be on the road;
\item The ego lane will be in the drivable area;
\item The sidewalk will be on either side of the road;
\item The pedestrians will either be on the sidewalk or the road;
\item The remainder of the environment surrounds the road.
\end{itemize}

Taking this domain knowledge into account, \tableref{tab:segclasses} introduces the \reducedmask{}, \wroadagentsmask{} and \wroadagentsnodrivablemask{} ordinal segmentation mask setups, including the ordinal relationship between classes in the form of a tree.  \figureref{fig:driving_scene1_bdd100k}(b) shows the \reducedmask{} ordinal segmentation mask setup for the autonomous driving scene in \figureref{fig:driving_scene1_bdd100k}(a).

\Figure{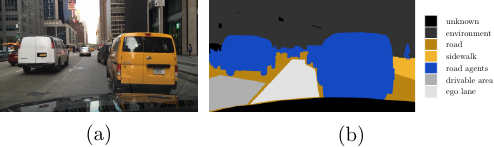}{\textbf{(a)~Driving scene from the BDD100K dataset~\cite{bdd100k}. (b)~Respective \reducedmask{} mask.}\label{fig:driving_scene1_bdd100k}}

\begin{table}
\centering
\caption{Segmentation classes. When using mask \reducedmask{}, the classes in \colorbox{black!25}{gray} are replaced by a single class ``road agents'' which is equal to their union. Mask \wroadagentsmask{} uses all of the shown classes, while mask \wroadagentsnodrivablemask{} does not use the drivable area task classes in \colorbox{blue!25}{blue}.}
\label{tab:segclasses}

\includegraphics{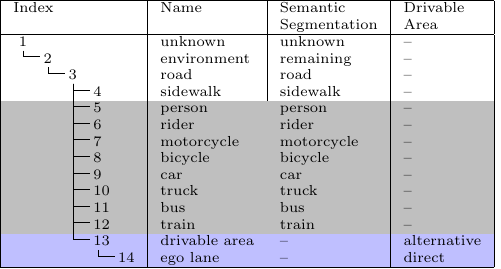}
\end{table}

\subsection{Experimental Setup} \label{section:results:setup}

Following the same procedure as \cite{fernandes_ordinal_2018}, experimental results were obtained using the UNet architecture~\cite{ronneberger_u-net_2015} with four groups of convolution blocks (each consisting of two convolution and one pooling layers) for each of the encoder and decoder portions\footnote{An open-source PyTorch implementation of the UNet architecture was used, \url{https://github.com/milesial/Pytorch-UNet}.}. All datasets were normalized with a mean of 0 and a standard deviation of 1 after data augmentation, consisting of random rotation, random horizontal flips, random crops, and random brightness and contrast. The networks were optimized for a maximum of 200 epochs in the case of the biomedical datasets and a maximum of 100 epochs in the case of the autonomous driving datasets, using the Adam optimizer with a learning rate of 1e-4 and a batch-size of 16. Early stopping was used with a patience of 15 epochs. After a train-test split of 80-20\%, a 5-fold training strategy consisting of 4 training folds and one validation fold was applied for each training dataset. For each fold, the best-performing model on the validation dataset was selected. An NVIDIA Tensor Core A100 GPU with 40GB of RAM and an NVIDIA RTX A2000 with 12GB of RAM were used to train the networks.

The metrics used were: (1)~the Dice coefficient, which evaluates the methods with respect to the ground truth labels, (2)~the contact surface, which evaluates the methods with respect to the structural ordinal consistency, and (3)~the percentage of unimodal pixels metrics, which evaluates the methods with respect to the pixel-wise ordinal consistency.

The cross-entropy loss and the methods by Fernandes \etal{}~\cite{fernandes_ordinal_2018} were used as the baseline models. The methods to be evaluated had their parameterization, including the range of regularization term weights ($\lambda$), empirically determined. These methods are:

\begin{itemize}
\item The semantic segmentation adaptation of the ordinal representation consistency $L_\text{O2}$ loss function, with the imposed margin $\delta = 0.05$, as recommended by the authors~\cite{albuquerque_ordinal_2021};
\item The proposed ordinal structural loss for segmentation, $L_\text{CSNP}$
and $L_\text{CSDT}$, with the distance transform threshold $\delta = 0.5$ and the maximum regularization distance $\gamma = 10$;
\item The mix of representation and structural methods, through $L_\text{CE}$ + $L_\text{O2}$ + $L_\text{CSNP}$ loss combination of the two regularization terms.
\end{itemize}


\section{Results}
\label{sec:results}

Tables~\ref{table:results-dice_coefficient_macro}--\ref{table:results-contact_surface} depict the main results. Since each loss term proposed is actually a penalty that is weighted with cross-entropy ($\text{loss}=\text{CE}+\lambda\,\text{term}$), each loss was evaluated using grid-search, $\lambda\in\{0.1,1,\dots,10^4\}$, selecting the best $\lambda$ according to the Dice metric. In the tables, bold indicates that the method has produced significantly better results than the baseline (just CE) using a one-tailed paired t-test with alpha=0.05.

Table~\ref{table:results-dice_coefficient_macro} shows that the proposed losses generally offer an improvement, as judged by the Dice coefficient. Notice that this metric is computed for each pixel, explaining why CE+O2 (a regularization term that focuses on the pixel-wise ordinal representation consistency) generally performs better. Two additional tables are presented with the proposed ordinal metrics, evaluating the consistency for the representation space (\% unimodality in Table~\ref{table:results-percentage_of_unimodal_px}) and the structural space (contact surface in Table~\ref{table:results-contact_surface}). Naturally, CE+O2 again shows a very good performance for representation consistency, but the other losses start to become competitive as the structural consistency is evaluated using the contact surface metric.

\begin{table*}
\caption{Overall results for the Dice coefficient metric (\%), a common metric for semantic segmentation. Introducing ordinal knowledge into the loss function improves this metric. Higher is better.}
\label{table:results-dice_coefficient_macro}
\begin{tabular}{|l|rrrrr|}
\hline
Dataset & CE & CE+O2 & CE+CSNP & CE+CSDT & CE+O2+CSNP \\\hline
Breast Aesthetics & $93.8\pm0.5$ & $94.2\pm0.6$ & $94.0\pm0.5$ & $94.0\pm0.4$ & $94.0\pm0.4$ \\
Cervix-MobileODT & $77.0\pm0.7$ & $77.2\pm0.4$ & $77.4\pm0.2$ & $76.9\pm0.7$ & $77.3\pm0.7$ \\
Mobbio & $93.8\pm0.1$ & $\mathbf{94.1\pm0.0}$ & $93.8\pm0.0$ & $93.7\pm0.2$ & $93.8\pm0.1$ \\
Teeth-ISBI & $74.0\pm1.6$ & $74.8\pm0.4$ & $74.9\pm1.4$ & $\mathbf{75.3\pm1.3}$ & $74.7\pm1.1$ \\
Teeth-UCV & $90.2\pm0.5$ & $90.8\pm0.4$ & $90.5\pm0.3$ & $90.3\pm0.3$ & $\mathbf{90.5\pm0.4}$ \\
\hline
BDDIntersected reduced & $71.2\pm0.5$ & $71.0\pm0.4$ & $71.2\pm0.6$ & $70.4\pm0.8$ & $71.9\pm0.2$ \\
BDDIntersected noabstract & $37.1\pm1.0$ & $\mathbf{38.6\pm1.0}$ & $38.1\pm0.7$ & $37.2\pm1.1$ & $\mathbf{38.5\pm1.1}$ \\
BDD10K & $38.1\pm1.1$ & $39.1\pm0.5$ & $38.2\pm1.2$ & $38.2\pm1.2$ & $39.2\pm0.6$ \\
Cityscapes & $36.0\pm1.3$ & $\mathbf{40.2\pm0.7}$ & $35.8\pm1.2$ & $36.5\pm1.4$ & $\mathbf{39.3\pm0.4}$ \\
\hline
\end{tabular}

\caption{Overall results for the percentage of unimodal pixels metric (\%). This is a representation metric that focuses on the ordinal consistency probabilities produced for each individual pixel. CE+O2, which also focuses on individual pixels, is better at addressing this metric. Higher is better.}
\label{table:results-percentage_of_unimodal_px}
\begin{tabular}{|l|rrrrr|}
\hline
Dataset & CE & CE+O2 & CE+CSNP & CE+CSDT & CE+O2+CSNP \\\hline
Breast Aesthetics & $6.2\pm0.7$ & $8.1\pm2.6$ & $5.9\pm2.1$ & $6.2\pm1.5$ & $6.7\pm0.9$ \\
Cervix-MobileODT & $1.0\pm0.2$ & $\mathbf{98.9\pm0.4}$ & $0.5\pm0.2$ & $0.9\pm0.3$ & $1.2\pm0.3$ \\
Mobbio & $0.8\pm0.1$ & $\mathbf{97.3\pm2.6}$ & $0.9\pm0.2$ & $0.7\pm0.2$ & $\mathbf{0.9\pm0.0}$ \\
Teeth-ISBI & $9.6\pm3.4$ & $\mathbf{67.1\pm3.8}$ & $\mathbf{11.2\pm4.6}$ & $\mathbf{13.5\pm3.4}$ & $\mathbf{15.5\pm4.4}$ \\
Teeth-UCV & $18.0\pm1.8$ & $\mathbf{96.8\pm1.0}$ & $25.7\pm6.8$ & $19.7\pm4.1$ & $\mathbf{73.3\pm6.6}$ \\
\hline
BDDIntersected reduced & $2.7\pm1.8$ & $3.4\pm2.0$ & $\mathbf{18.8\pm9.4}$ & $1.6\pm1.3$ & $6.4\pm4.0$ \\
BDDIntersected noabstract & $2.4\pm1.9$ & $\mathbf{11.7\pm4.1}$ & $3.7\pm2.8$ & $6.8\pm5.2$ & $\mathbf{12.7\pm6.1}$ \\
BDD10K & $32.3\pm6.4$ & $36.1\pm9.2$ & $31.7\pm1.7$ & $34.4\pm2.7$ & $\mathbf{49.9\pm9.3}$ \\
Cityscapes & $23.0\pm12.0$ & $\mathbf{69.9\pm5.4}$ & $38.1\pm3.9$ & $24.9\pm2.4$ & $\mathbf{88.4\pm4.0}$ \\
\hline
\end{tabular}

\caption{Overall results for the contact surface metric (\%). This is a structural metric that measures consistency between adjacent pixels. In this case, the proposed terms (CSNP and CSDT) improve the ordinal differentiation between neighbors. Lower is better.}
\label{table:results-contact_surface}
\begin{tabular}{|l|rrrrr|}
\hline
Dataset & CE & CE+O2 & CE+CSNP & CE+CSDT & CE+O2+CSNP \\\hline
Breast Aesthetics & $0.2\pm0.2$ & $0.2\pm0.2$ & $0.1\pm0.1$ & $0.2\pm0.2$ & $0.3\pm0.2$ \\
Cervix-MobileODT & $14.5\pm3.3$ & $\mathbf{1.1\pm0.4}$ & $10.8\pm2.7$ & $13.6\pm1.7$ & $14.5\pm2.5$ \\
Mobbio & $12.3\pm0.5$ & $\mathbf{4.2\pm0.2}$ & $12.3\pm0.4$ & $12.4\pm0.6$ & $\mathbf{11.5\pm0.3}$ \\
Teeth-ISBI & $30.0\pm3.9$ & $26.1\pm2.4$ & $28.4\pm1.4$ & $27.1\pm1.8$ & $27.5\pm1.2$ \\
Teeth-UCV & $7.0\pm1.4$ & $\mathbf{3.0\pm0.7}$ & $\mathbf{2.3\pm1.0}$ & $5.5\pm1.0$ & $\mathbf{1.8\pm0.5}$ \\
\hline
BDDIntersected reduced & $55.9\pm4.0$ & $54.9\pm2.9$ & $\mathbf{49.1\pm2.3}$ & $54.5\pm3.8$ & $52.8\pm3.0$ \\
BDDIntersected noabstract & $47.8\pm2.6$ & $47.8\pm2.1$ & $47.8\pm2.6$ & $46.6\pm1.6$ & $47.9\pm2.0$ \\
BDD10K & $51.2\pm2.2$ & $50.5\pm2.8$ & $50.3\pm1.4$ & $50.9\pm1.5$ & $\mathbf{48.4\pm2.3}$ \\
Cityscapes & $59.3\pm2.2$ & $61.4\pm1.8$ & $56.9\pm3.7$ & $60.2\pm2.8$ & $\mathbf{55.5\pm2.2}$ \\
\hline
\end{tabular}

\bigskip
(As \textbf{bold} when the method is better than the baseline (CE) using a one-tailed paired t-test with alpha=0.05.)
\end{table*}

\Figure{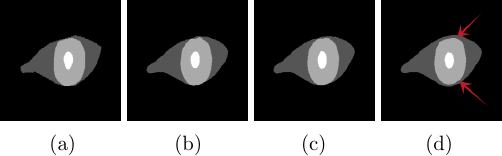}{\textbf{Comparison of the effect of changing the $\lambda$ coefficient controlling the strength of the CSDT loss term (Mobbio dataset) with (a)~Ground-truth, (b)~$\lambda=0.1$, (c)~$\lambda=1$, (d)~$\lambda=10$. The latter (d) is an example of excessive regularization, where the model includes a sclera border (dark gray) between the background (black) and the iris (light gray), which does not exist in the ground truth.}\label{fig:results:bio}}

Additional results are shown in plots that depict the influence of the weight of the regularization term, $\lambda$, on each respective metric. Figures \figurerefnotext{fig:biomedical:plots:dice}, \figurerefnotext{fig:biomedical:plots:cs} and \figurerefnotext{fig:biomedical:plots:unimodal} show, respectively, the Dice coefficient, contact surface, and unimodal metrics for the biomedical datasets. We can see that the ordinal methods successfully optimize the ordinal metrics, resulting in more ordinally consistent models. The dice coefficient metric does not improve much; however, excessive regularization can occur at some $\lambda$ values, resulting in a lower Dice metric. An example of this over-regularization is presented in~\figureref{fig:results:bio}.

\figureref{fig:bdd100k:plots:dice} shows the Dice coefficient for the autonomous driving datasets, including the out-of-distribution domain testing. There, it can be seen that CE+O2 is the method with the highest impact on the generalization capability of the model, achieving a maximum absolute gain of $4.2\%$ ($11.5\%$ in relative terms) at $\lambda = 10$, which means that using this loss helps the model generalize better to previously unseen scenarios. \figureref{fig:results:co2:samples} showcases a set of model inference outputs for this scenario, comparing cross-entropy outputs to CE+$\lambda$\,O2 with $\lambda = 10$.

To evaluate whether the inclusion of domain knowledge during the training of the models would help the network learn better with scarce data for autonomous driving scenarios, an experiment that consisted of varying the scale of the dataset used to train the models was performed. For each method, the best-performing lambda in the out-of-distribution scenario was selected, the rationale being that those models are the best at generalizing to unseen scenarios, which is useful when training with low amounts of data. For the $L_\text{O2}$ and $L_\text{CSNP}$ terms, $\lambda = 10$ was used, and for the $L_\text{CSDT}$, $\lambda = 0.1$ was used. \figureref{fig:bdd100k:plots:scalevar:dice} shows the Dice coefficient results for the dataset scale variation experiment.

When testing with the BDD10K dataset, CE+O2 and CE+CSNP regularization terms achieve better Dice coefficient results at scales $0.25$ and $0.5$ when compared with the CE baseline, which suggests that, indeed, these losses help the network learn better when data is scarce. Especially when using the CE+O2 loss, resulting in absolute gains of $1.2\%$ ($5.7\%$ in relative terms) in the Dice coefficient performance at scale $0.25$ over CE. CE+O2 continues beating CE Dice performance through scales $0.1$ and $0.05$.

Cross-entropy alone generalizes better than when used in conjunction with the ordinal regularizers when testing with the Cityscapes dataset in an out-of-distribution scenario for lower scales. However, CE+O2 continues to generalize better than the cross-entropy throughout all scales, achieving a maximum absolute gain of $5.3\%$ ($15.7\%$ in relative terms) in the Dice coefficient at scale $0.75$. Still, it can be seen that the generalization ability of CE+O2 decreases at a higher rate than its performance on BDD10K -- training with fewer data has a higher impact on the model's generalization ability.

Regarding the ordinal metrics, the scale variance does not significantly affect either metric.

In terms of compute complexity, since these are different losses, they do not affect inference time. Furthermore, any change in training time was too small to be detected.

\Figure[clip, trim=0.6cm 1cm 1.8cm 1.65cm, width=8.5cm]{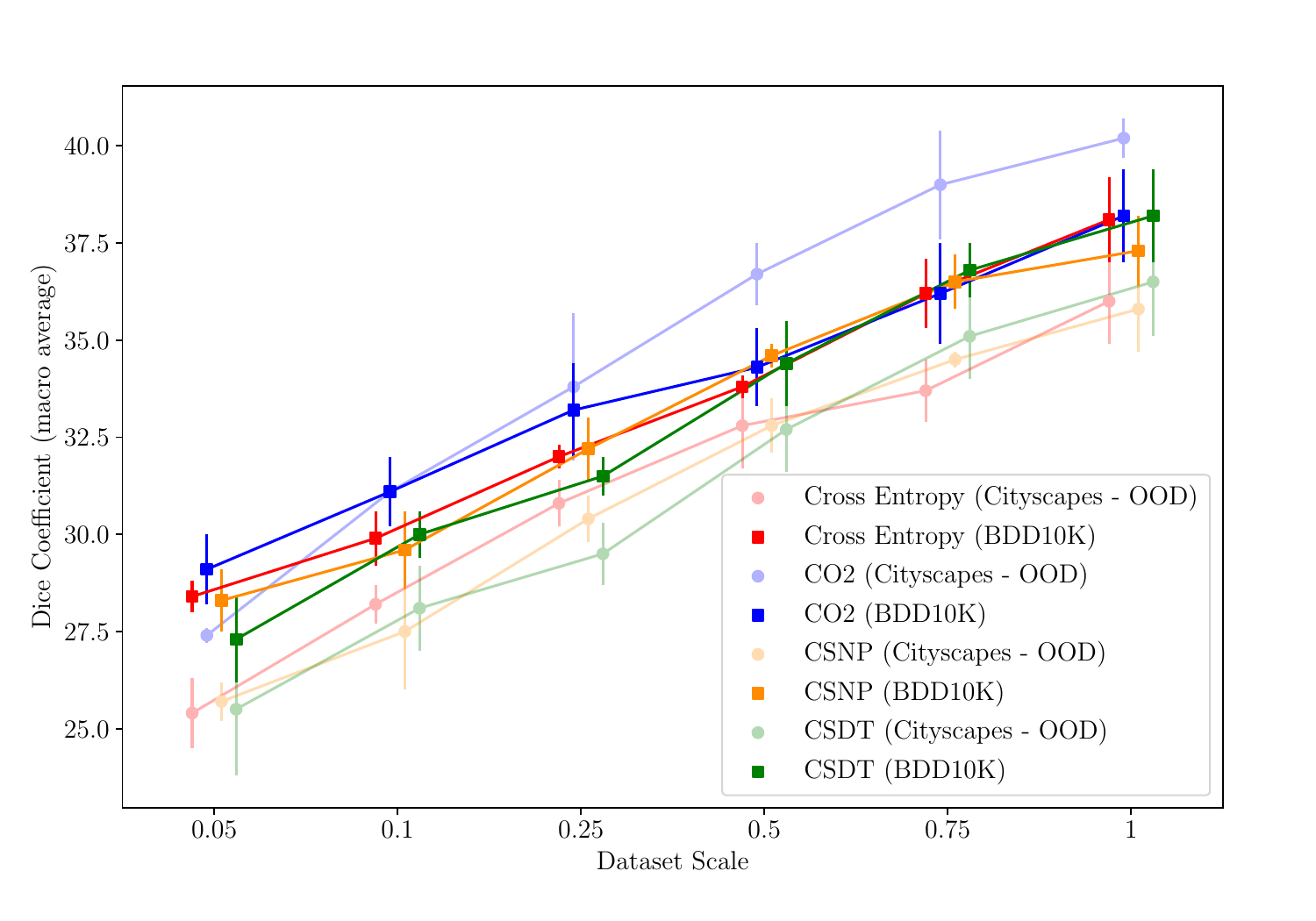}{\textbf{Dice coefficient (macro average) results for the autonomous driving datasets scale variation experiments (higher is better).}\label{fig:bdd100k:plots:scalevar:dice}}

\subsection{Discussion}

\Figure[clip, trim=0.9cm 0.8cm 0.8cm 0.7cm, width=8.5cm]{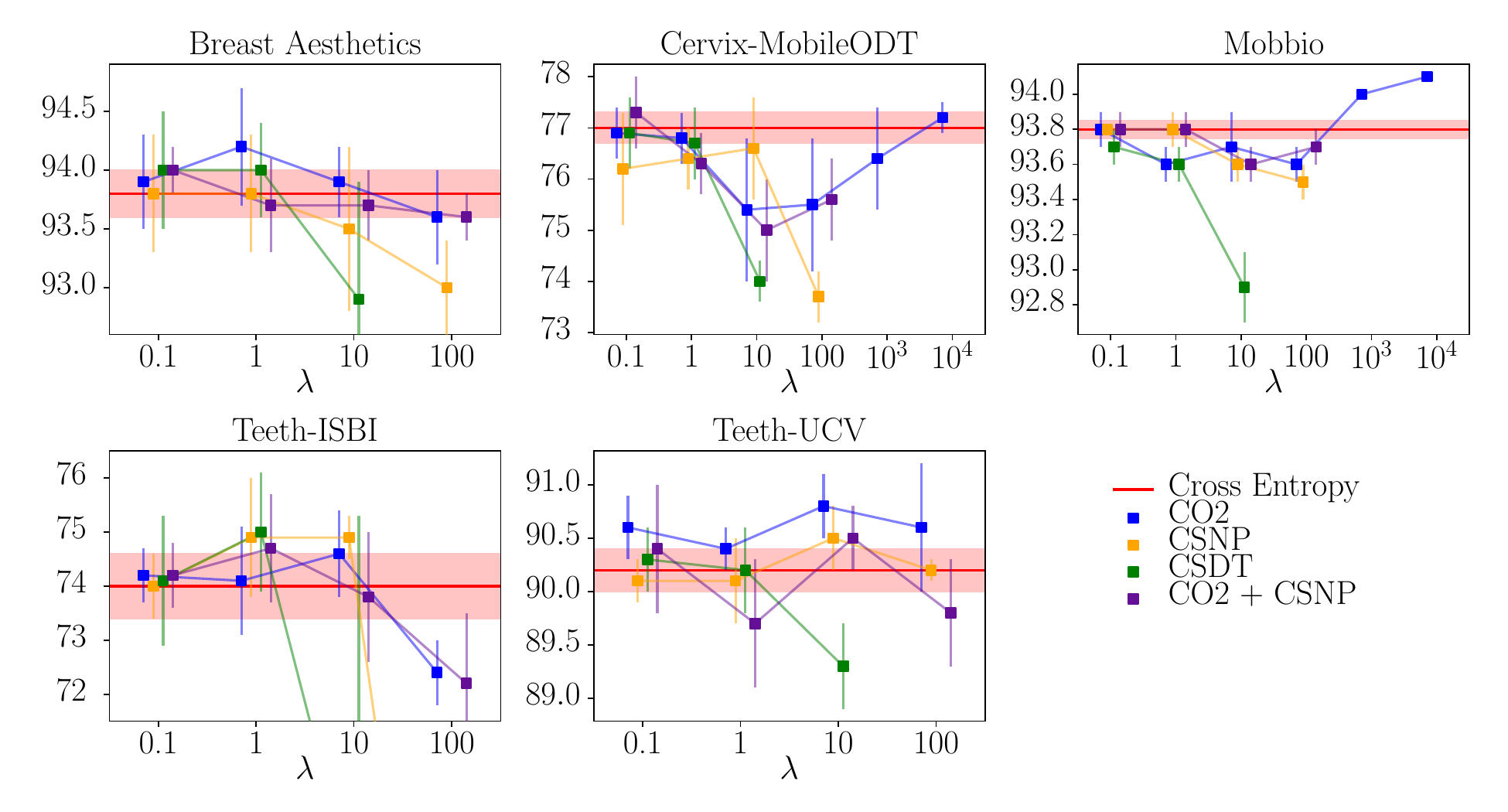}{\textbf{Dice coefficient (macro average) results for the biomedical datasets (higher is better).}\label{fig:biomedical:plots:dice}}

\Figure[clip, trim=0.9cm 0.8cm 0.8cm 0.7cm, width=8.5cm]{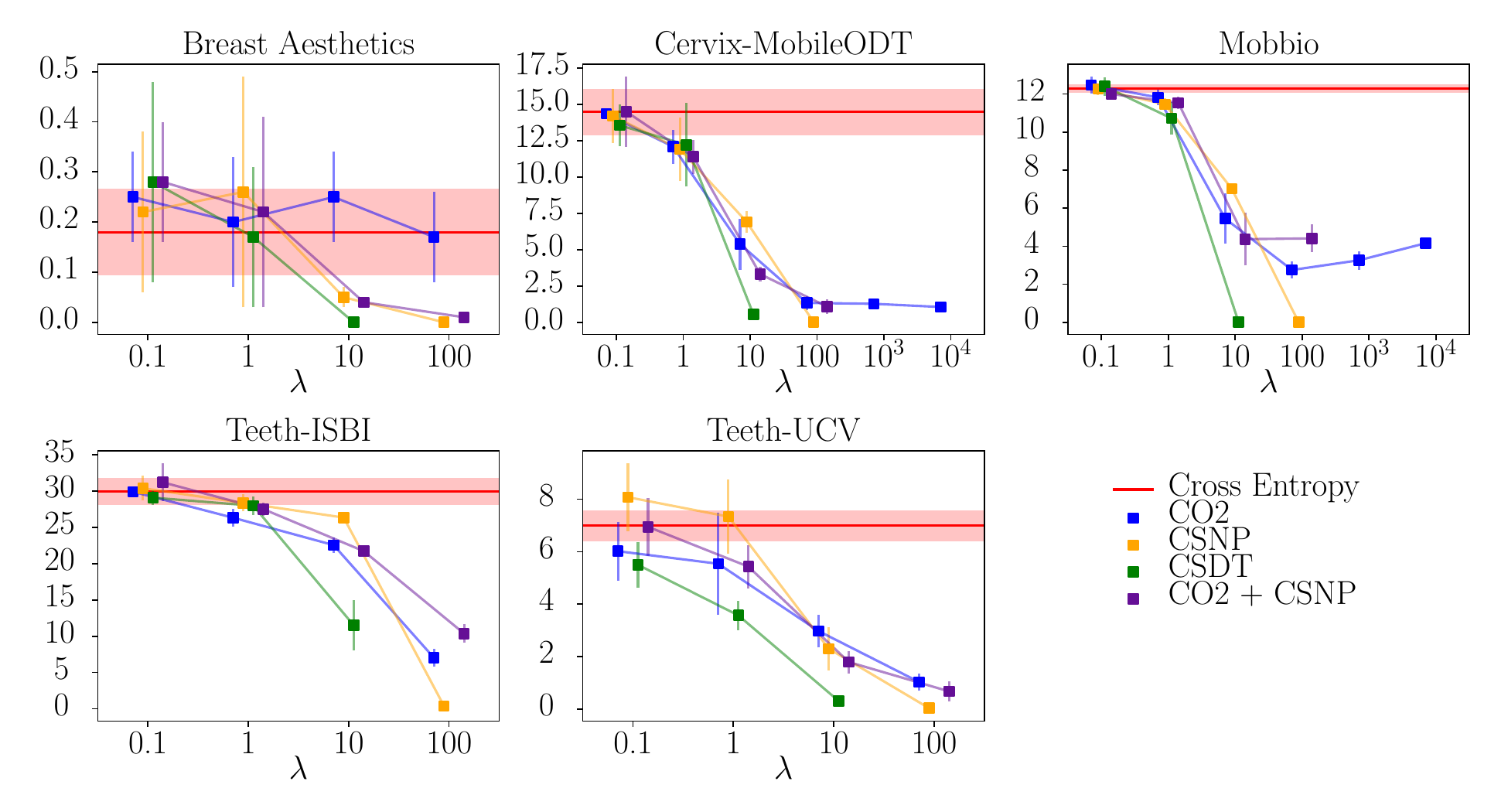}{\textbf{Contact surface results for the biomedical datasets (lower is better).}\label{fig:biomedical:plots:cs}}

\Figure[clip, trim=0.9cm 0.8cm 0.8cm 0.7cm, width=8.5cm]{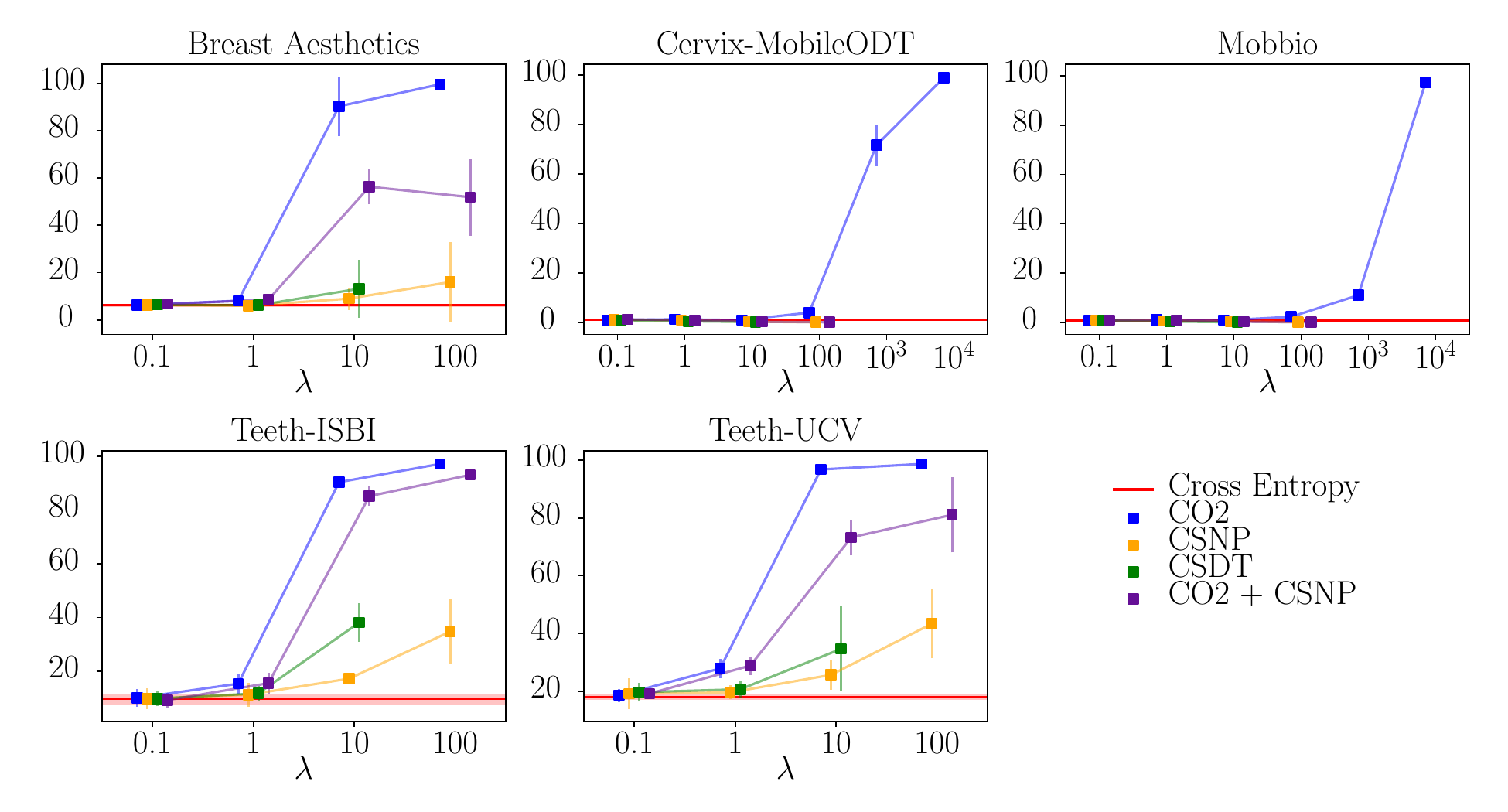}{\textbf{Percentage of unimodal pixels results for the biomedical datasets (higher is better).}\label{fig:biomedical:plots:unimodal}}

\Figure[clip, trim=1.4cm 1.1cm 8.5cm 1.15cm, width=8.5cm]{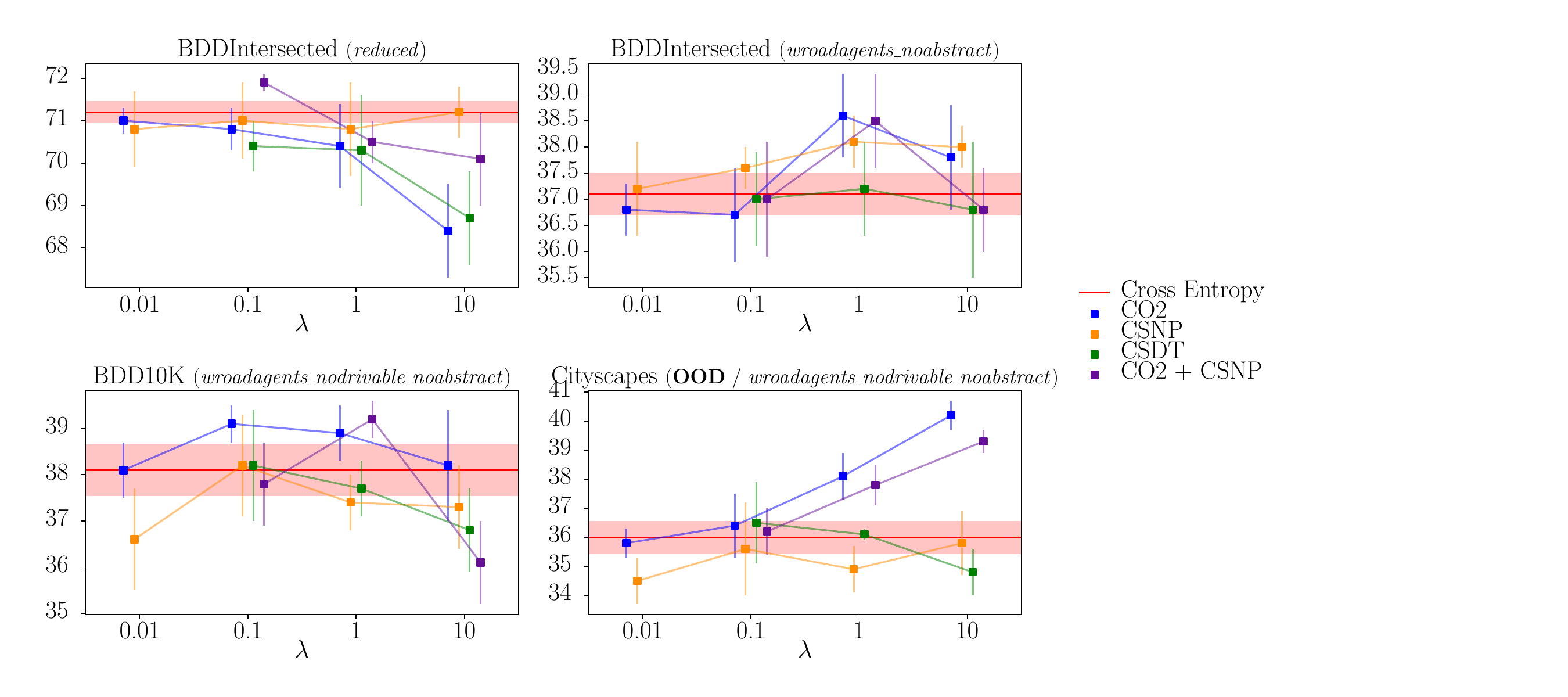}{\textbf{Dice coefficient (macro average) results for the autonomous driving datasets (higher is better).}\label{fig:bdd100k:plots:dice}}

\Figure{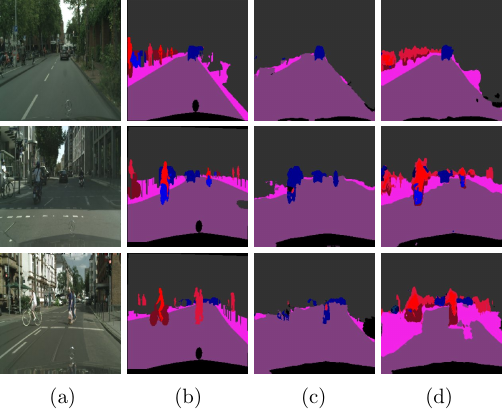}{\textbf{Comparison of the influence of cross-entropy and O2 term on model output in out-of-distribution inference. (a)~Input image. (b)~Ground-truth. (c)~CE alone. (d)~CE+O2 ($\lambda = 10.0$). The O2 term promotes more accurate identification of pedestrians, two-wheel vehicles, and riders when compared with the cross-entropy loss output.}\label{fig:results:co2:samples}}

A comparison between the biomedical and autonomous driving datasets results concludes that the biomedical results are significantly better in terms of the Dice coefficient performance -- the performance gains with the autonomous driving datasets are still positive, but not as large, since these datasets are considerably more complex, with a greater variety of scenarios and segmentation classes.

The representation space, \ie pixel-wise, $L_\text{O2}$ performed well in both the biomedical and autonomous driving datasets, especially when considered in an out-of-distribution domain -- it could potentially be applied in a real-world scenario in order to improve the generalization capabilities of perception algorithms.

The structured space regularization terms, $L_\text{CSNP}$ and $L_\text{CSDT}$, may not apply to the autonomous driving scenario in their current form, at least at high regularization weights. The scene perspective from the car makes it so that there is a large amount of valid contact surface between non-ordinally adjacent classes -- in a 2D projection of the real world, the absolute minimization of these contact surfaces may not be the best solution since occlusions and different perspectives may originate legitimate contact between non-ordinally adjacent classes. Relaxed adaptations of these methods that consider this type of contact could be devised in the future. However, in the biomedical datasets, these methods performed well, with their greatest difficulty being the existence of occlusions.

As seen, the choice of $\lambda$ value, \ie regularization weight, is critical for the performance of the proposed methods. For this reason, in order to be applied to different domains, there should be an empirical study of the influence of the lambda value in the segmentation performance in that specific domain and the choice of a value that is a balance between the ordinal metrics and the Dice coefficient, \ie a value that promotes some unimodality and structural consistency but also does not hurt Dice performance to the point where it is unusable.


\section{Conclusion}
\label{sec:conclusion}

This paper addresses the overlooked role of ordinality in semantic segmentation by introducing methods that enforce both pixel-level and structural ordinal consistency. By leveraging the inherent relationships between classes, we developed two regularization loss functions and an evaluation metric tailored to promote ordinal consistency in structured image data.

In this domain, two categories of loss functions for ordinal segmentation were studied: (1)~an ordinal representation consistency loss, where each pixel is treated individually by promoting unimodality in its probability distribution, and (2)~a structural consistency loss, where each pixel is considered in the context of its neighborhood and the contact surface between non-ordinally adjacent classes is minimized.

For that purpose, the following loss terms were proposed:
\begin{enumerate}
\item a segmentation adaptation of the representation $L_\text{O2}$ loss which penalizes ordinal inconsistencies within each pixel;
\item the structural $L_\text{CSNP}$ loss, which considers only the immediate neighbor pixels;
\item the structural $L_\text{CSDT}$ loss, which considers the global neighborhood.
\end{enumerate}
In addition, two metrics were proposed to evaluate the network output's ordinal consistency: (1)~the percentage of unimodal pixels and (2)~the contact surface between the segmentation masks of non-ordinally adjacent classes. The first evaluates in-pixel consistency, while the second evaluates spatial consistency.

The proposed methods were initially validated on five biomedical datasets and two autonomous driving datasets, resulting in more ordinally consistent models without significantly impacting the Dice coefficient. The resulting models were tested in an autonomous driving out-of-distribution scenario to evaluate the methods' impact on the models' generalization capability. Furthermore, the autonomous driving models were also tested with scaled-down versions of the BDD100K dataset to evaluate how the network learns with scarce data. The ordinal methods achieved maximum improvements in the Dice coefficient with an absolute value of $5.3\%$ ($15.7\%$ in relative terms) in the out-of-distribution domain.

To summarize, incorporating ordinal consistency into semantic segmentation models showed promising results, including developing more generalizable models that exhibit improved learning capabilities with limited data availability. Since these are loss terms, they do not add time complexity to inference time.

Future research topics could include: (i)~the development of more flexible structural ordinal segmentation methods, allowing for limited contact between non-ordinally adjacent classes, such as in the case of occlusions and different perspectives; and (ii)~the development of novel methods that leverage ordinal constraints not necessarily consisting of augmented loss functions.

Furthermore, the current work could contribute to other types of consistency. The partially ordered domains explored in this work were structural (e.g., a vehicle may be located on the road or in parking spaces), but a representational partially ordered domain could correspond to hierarchical segmentation~\cite{li2022deep}, where each pixel has a taxonomy (e.g., a motorcycle and bicycle are both two-wheels; two-wheels and four-wheels are both types of vehicles).
Also, part segmentation~\cite{bo2011shape,tsogkas2015deep} could be considered a different type of structural consistency, where parts of an object are next to each other (e.g., the ``head'' appearing above ``upper body'' or ``hair'' above ``head'').

\appendix
\section{Source Code}
The full source code is provided in \url{https://github.com/rafaavc/ordinal-semantic-segmentation}. The code for the proposed spatial losses is small and is presented here in full:

{\scriptsize
\begin{verbatim}
def CSNP(P, K):
    loss = 0
    count = 0

    # for each pair of non-ordinally adjacent classes
    for k1 in range(K):
        for k2 in range(K):
            if abs(k2 - k1) <= 1:
                continue
            
            # more weight to more ordinally distant classes
            ordinal_multiplier = abs(k2 - k1) - 1

            dx = P[:,  k1, :, :-1] * P[:, k2, :, 1:]
            dy = P[:,  k1, :-1, :] * P[:, k2, 1:, :]

            loss += ordinal_multiplier * \
                (torch.mean(dx) + torch.mean(dy))/2
            count += 1
    
    if count != 0:
        loss /= count
    return loss
\end{verbatim}

\begin{verbatim}
def CSDT(P, K, threshold=.5, max_dist=10.):
    loss = 0
    count = 0
    
    activations = 1. * (P > threshold)
    DT = distance_transform(activations)

    # cap the maximum distance at 10
    max_dist_DT = (DT >= max_dist) * max_dist
    # select the values with a distance < 10
    DT *= DT < max_dist
    # add the capped values
    DT += max_dist_DT

    # for each pair of non-ordinally adjacent classes
    for k1 in range(K):
        for k2 in range(k1 + 2, K):
            # more weight to more ordinally distant classes
            ordinal_multiplier = abs(k2 - k1) - 1

            d_k1, d_k2 = DT[:, k1], DT[:, k2]
            p_k1, p_k2 = P[:, k1], P[:, k2]

            calc = p_k1 * d_k2 + p_k2 * d_k1
            calc = calc[calc != 0]

            loss += ordinal_multiplier * torch.mean(calc)
            count += 1

    if count != 0:
        loss /= count
    loss /= max_dist  # normalize
    return -loss  # maximize
\end{verbatim}}

\bibliographystyle{IEEEtran}
\bibliography{refs}

\begin{thebibliography}{10}
\providecommand{\url}[1]{#1}
\csname url@samestyle\endcsname
\providecommand{\newblock}{\relax}
\providecommand{\bibinfo}[2]{#2}
\providecommand{\BIBentrySTDinterwordspacing}{\spaceskip=0pt\relax}
\providecommand{\BIBentryALTinterwordstretchfactor}{4}
\providecommand{\BIBentryALTinterwordspacing}{\spaceskip=\fontdimen2\font plus
\BIBentryALTinterwordstretchfactor\fontdimen3\font minus
  \fontdimen4\font\relax}
\providecommand{\BIBforeignlanguage}[2]{{%
\expandafter\ifx\csname l@#1\endcsname\relax
\typeout{** WARNING: IEEEtran.bst: No hyphenation pattern has been}%
\typeout{** loaded for the language `#1'. Using the pattern for}%
\typeout{** the default language instead.}%
\else
\language=\csname l@#1\endcsname
\fi
#2}}
\providecommand{\BIBdecl}{\relax}
\BIBdecl

\bibitem{zakaria2023lane}
N.~J. Zakaria, M.~I. Shapiai, R.~Abd~Ghani, M.~N.~M. Yassin, M.~Z. Ibrahim, and
  N.~Wahid, ``Lane detection in autonomous vehicles: A systematic review,''
  \emph{IEEE Access}, vol.~11, pp. 3729--3765, 2023.

\bibitem{yousri2021deep}
R.~Yousri, M.~A. Elattar, and M.~S. Darweesh, ``A deep learning-based
  benchmarking framework for lane segmentation in the complex and dynamic road
  scenes,'' \emph{IEEE Access}, vol.~9, pp. 117\,565--117\,580, 2021.

\bibitem{zeiler2014visualizing}
M.~D. Zeiler and R.~Fergus, ``Visualizing and understanding convolutional
  networks,'' in \emph{Computer Vision--ECCV 2014: 13th European Conference,
  Zurich, Switzerland, September 6-12, 2014, Proceedings, Part I 13}.\hskip 1em
  plus 0.5em minus 0.4em\relax Springer, 2014, pp. 818--833.

\bibitem{cruz2017ordinal}
R.~Cruz, K.~Fernandes, J.~F. Pinto~Costa, M.~P. Ortiz, and J.~S. Cardoso,
  ``Ordinal class imbalance with ranking,'' in \emph{Pattern Recognition and
  Image Analysis: 8th Iberian Conference, IbPRIA 2017, Faro, Portugal, June
  20-23, 2017, Proceedings 8}.\hskip 1em plus 0.5em minus 0.4em\relax Springer,
  2017, pp. 3--12.

\bibitem{albuquerque_ordinal_2021}
\BIBentryALTinterwordspacing
T.~Albuquerque, R.~Cruz, and J.~S. Cardoso, ``\BIBforeignlanguage{en}{Ordinal
  losses for classification of cervical cancer risk},''
  \emph{\BIBforeignlanguage{en}{PeerJ Computer Science}}, vol.~7, p. e457, Apr.
  2021, publisher: PeerJ Inc. [Online]. Available:
  \url{https://peerj.com/articles/cs-457}
\BIBentrySTDinterwordspacing

\bibitem{albuquerque2022quasi}
------, ``Quasi-unimodal distributions for ordinal classification,''
  \emph{Mathematics}, vol.~10, no.~6, p. 980, 2022.

\bibitem{bellmann2020ordinal}
P.~Bellmann and F.~Schwenker, ``Ordinal classification: Working definition and
  detection of ordinal structures,'' \emph{IEEE Access}, vol.~8, pp.
  164\,380--164\,391, 2020.

\bibitem{fu2018deep}
H.~Fu, M.~Gong, C.~Wang, K.~Batmanghelich, and D.~Tao, ``Deep ordinal
  regression network for monocular depth estimation,'' in \emph{Proceedings of
  the IEEE conference on computer vision and pattern recognition}, 2018, pp.
  2002--2011.

\bibitem{fernandes_ordinal_2018}
K.~Fernandes and J.~S. Cardoso, ``Ordinal image segmentation using deep neural
  networks,'' in \emph{2018 {International} {Joint} {Conference} on {Neural}
  {Networks} ({IJCNN})}, Jul. 2018, pp. 1--7, iSSN: 2161-4407.

\bibitem{melo2023retinal}
T.~Melo, {\^A}.~Carneiro, A.~Campilho, and A.~M. Mendon{\c{c}}a, ``Retinal
  layer and fluid segmentation in optical coherence tomography images using a
  hierarchical framework,'' \emph{Journal of Medical Imaging}, vol.~10, no.~1,
  pp. 014\,006--014\,006, 2023.

\bibitem{li2022deep}
L.~Li, T.~Zhou, W.~Wang, J.~Li, and Y.~Yang, ``Deep hierarchical semantic
  segmentation,'' in \emph{Proceedings of the IEEE/CVF Conference on Computer
  Vision and Pattern Recognition}, 2022, pp. 1246--1257.

\bibitem{bo2011shape}
Y.~Bo and C.~C. Fowlkes, ``Shape-based pedestrian parsing,'' in \emph{CVPR
  2011}.\hskip 1em plus 0.5em minus 0.4em\relax IEEE, 2011, pp. 2265--2272.

\bibitem{tsogkas2015deep}
S.~Tsogkas, I.~Kokkinos, G.~Papandreou, and A.~Vedaldi, ``Deep learning for
  semantic part segmentation with high-level guidance,'' \emph{arXiv preprint
  arXiv:1505.02438}, 2015.

\bibitem{zhao2019ordinal}
Y.~Zhao, J.~Li, Y.~Zhang, Y.~Song, and Y.~Tian, ``Ordinal multi-task part
  segmentation with recurrent prior generation,'' \emph{IEEE transactions on
  pattern analysis and machine intelligence}, vol.~43, no.~5, pp. 1636--1648,
  2019.

\bibitem{cristino2023introducing}
R.~V. Cristino, ``Introducing domain knowledge to scene parsing in autonomous
  driving,'' Master's thesis, University of Porto, 2023.

\bibitem{long2015fully}
J.~Long, E.~Shelhamer, and T.~Darrell, ``Fully convolutional networks for
  semantic segmentation,'' in \emph{Proceedings of the IEEE Conference on
  Computer Vision and Pattern Recognition}, 2015, pp. 3431--3440.

\bibitem{ronneberger_u-net_2015}
\BIBentryALTinterwordspacing
O.~Ronneberger, P.~Fischer, and T.~Brox, ``U-{Net}: {Convolutional} {Networks}
  for {Biomedical} {Image} {Segmentation},'' May 2015, arXiv:1505.04597 [cs]
  version: 1. [Online]. Available: \url{http://arxiv.org/abs/1505.04597}
\BIBentrySTDinterwordspacing

\bibitem{badrinarayanan2017segnet}
V.~Badrinarayanan, A.~Kendall, and R.~Cipolla, ``{SegNet}: A deep convolutional
  encoder-decoder architecture for image segmentation,'' \emph{IEEE
  Transactions on Pattern Analysis and Machine Intelligence}, vol.~39, no.~12,
  pp. 2481--2495, 2017.

\bibitem{chen2018encoder}
L.-C. Chen, Y.~Zhu, G.~Papandreou, F.~Schroff, and H.~Adam, ``Encoder-decoder
  with atrous separable convolution for semantic image segmentation,'' in
  \emph{Proceedings of the European conference on computer vision (ECCV)},
  2018, pp. 801--818.

\bibitem{das2021estimation}
S.~Das, A.~A. Fime, N.~Siddique, and M.~Hashem, ``Estimation of road boundary
  for intelligent vehicles based on {DeepLabv3+} architecture,'' \emph{IEEE
  Access}, vol.~9, pp. 121\,060--121\,075, 2021.

\bibitem{zhao2017pyramid}
H.~Zhao, J.~Shi, X.~Qi, X.~Wang, and J.~Jia, ``Pyramid scene parsing network,''
  in \emph{Proceedings of the IEEE conference on computer vision and pattern
  recognition}, 2017, pp. 2881--2890.

\bibitem{zongren2023densetrans}
L.~ZongRen, W.~Silamu, W.~Yuzhen, and W.~Zhe, ``{DenseTrans}: {Multimodal}
  brain tumor segmentation using {Swin} transformer,'' \emph{IEEE Access},
  vol.~11, pp. 42\,895--42\,908, 2023.

\bibitem{pinto_da_costa_unimodal_2008}
\BIBentryALTinterwordspacing
J.~F. Pinto~da Costa, H.~Alonso, and J.~S. Cardoso,
  ``\BIBforeignlanguage{en}{The unimodal model for the classification of
  ordinal data},'' \emph{\BIBforeignlanguage{en}{Neural Networks}}, vol.~21,
  no.~1, pp. 78--91, Jan. 2008. [Online]. Available:
  \url{https://www.sciencedirect.com/science/article/pii/S089360800700202X}
\BIBentrySTDinterwordspacing

\bibitem{beckham_unimodal_2017}
\BIBentryALTinterwordspacing
C.~Beckham and C.~Pal, ``Unimodal probability distributions for deep ordinal
  classification,'' Jun. 2017, arXiv:1705.05278 [stat]. [Online]. Available:
  \url{http://arxiv.org/abs/1705.05278}
\BIBentrySTDinterwordspacing

\bibitem{cardoso_unimodal_2023}
\BIBentryALTinterwordspacing
J.~S. Cardoso, R.~Cruz, and T.~Albuquerque, ``Unimodal {Distributions} for
  {Ordinal} {Regression},'' Mar. 2023, arXiv:2303.04547 [cs]. [Online].
  Available: \url{http://arxiv.org/abs/2303.04547}
\BIBentrySTDinterwordspacing

\bibitem{cheng_neural_2007}
\BIBentryALTinterwordspacing
J.~Cheng, ``A neural network approach to ordinal regression,'' Apr. 2007,
  arXiv:0704.1028 [cs]. [Online]. Available:
  \url{http://arxiv.org/abs/0704.1028}
\BIBentrySTDinterwordspacing

\bibitem{cardoso2010classification}
J.~S. Cardoso and R.~Sousa, ``Classification models with global constraints for
  ordinal data,'' in \emph{2010 Ninth International Conference on Machine
  Learning and Applications}.\hskip 1em plus 0.5em minus 0.4em\relax IEEE,
  2010, pp. 71--77.

\bibitem{sousa2011ensemble}
R.~Sousa and J.~S. Cardoso, ``Ensemble of decision trees with global
  constraints for ordinal classification,'' in \emph{2011 11th International
  Conference on Intelligent Systems Design and Applications}.\hskip 1em plus
  0.5em minus 0.4em\relax IEEE, 2011, pp. 1164--1169.

\bibitem{strutz2021distance}
T.~Strutz, ``The distance transform and its computation,'' \emph{arXiv preprint
  arXiv:2106.03503}, 2021.

\bibitem{cardoso_towards_2007}
\BIBentryALTinterwordspacing
J.~S. Cardoso and M.~J. Cardoso, ``\BIBforeignlanguage{en}{Towards an
  intelligent medical system for the aesthetic evaluation of breast cancer
  conservative treatment},'' \emph{\BIBforeignlanguage{en}{Artificial
  Intelligence in Medicine}}, vol.~40, no.~2, pp. 115--126, Jun. 2007.
  [Online]. Available:
  \url{https://www.sciencedirect.com/science/article/pii/S0933365707000206}
\BIBentrySTDinterwordspacing

\bibitem{noauthor_intel_nodate}
\BIBentryALTinterwordspacing
Intel, ``\BIBforeignlanguage{en}{Intel \& {MobileODT} {Cervical} {Cancer}
  {Screening}},''
  \url{https://kaggle.com/competitions/intel-mobileodt-cervical-cancer-screening},
  accessed: 2023-06-22. [Online]. Available:
  \url{https://kaggle.com/competitions/intel-mobileodt-cervical-cancer-screening}
\BIBentrySTDinterwordspacing

\bibitem{sequeira_mobbio_2014}
A.~F. Sequeira, J.~C. Monteiro, A.~Rebelo, and H.~P. Oliveira, ``{MobBIO}: {A}
  multimodal database captured with a portable handheld device,'' in \emph{2014
  {International} {Conference} on {Computer} {Vision} {Theory} and
  {Applications} ({VISAPP})}, vol.~3, Jan. 2014, pp. 133--139.

\bibitem{wang_benchmark_2016}
\BIBentryALTinterwordspacing
C.-W. Wang, C.-T. Huang, J.-H. Lee, C.-H. Li, S.-W. Chang, M.-J. Siao, T.-M.
  Lai, B.~Ibragimov, T.~Vrtovec, O.~Ronneberger, P.~Fischer, T.~F. Cootes, and
  C.~Lindner, ``\BIBforeignlanguage{en}{A benchmark for comparison of dental
  radiography analysis algorithms},'' \emph{\BIBforeignlanguage{en}{Medical
  Image Analysis}}, vol.~31, pp. 63--76, Jul. 2016. [Online]. Available:
  \url{https://www.sciencedirect.com/science/article/pii/S1361841516000190}
\BIBentrySTDinterwordspacing

\bibitem{fernandez_teethpalate_2012}
K.~Fernandez and C.~Chang, ``\BIBforeignlanguage{en}{Teeth/{Palate} and
  {Interdental} {Segmentation} {Using} {Artificial} {Neural} {Networks}},'' in
  \emph{\BIBforeignlanguage{en}{Artificial {Neural} {Networks} in {Pattern}
  {Recognition}}}, ser. Lecture {Notes} in {Computer} {Science}, N.~Mana,
  F.~Schwenker, and E.~Trentin, Eds.\hskip 1em plus 0.5em minus 0.4em\relax
  Berlin, Heidelberg: Springer, 2012, pp. 175--185.

\bibitem{bdd100k}
F.~Yu, H.~Chen, X.~Wang, W.~Xian, Y.~Chen, F.~Liu, V.~Madhavan, and T.~Darrell,
  ``{BDD100K}: A diverse driving dataset for heterogeneous multitask
  learning,'' in \emph{IEEE/CVF Conference on Computer Vision and Pattern
  Recognition (CVPR)}, June 2020.

\bibitem{cordts_cityscapes_2016}
\BIBentryALTinterwordspacing
M.~Cordts, M.~Omran, S.~Ramos, T.~Rehfeld, M.~Enzweiler, R.~Benenson,
  U.~Franke, S.~Roth, and B.~Schiele, ``The {Cityscapes} {Dataset} for
  {Semantic} {Urban} {Scene} {Understanding},'' Apr. 2016, arXiv:1604.01685
  [cs]. [Online]. Available: \url{http://arxiv.org/abs/1604.01685}
\BIBentrySTDinterwordspacing

\end{thebibliography}

\end{document}